\documentclass[lettarpaper, 10 pt, conference]{ieeeconf}  % Comment this line out if you need a4paper

\IEEEoverridecommandlockouts                              % This command is only needed if 
\usepackage{graphicx} % for pdf, bitmapped graphics files

\newcommand{\argmin}{\mathop{\rm arg~min}\limits}

\title{\LARGE \bf
	LiDAR and Camera Calibration using Motion Estimated by Sensor Fusion Odometry
}

\author{Ryoichi Ishikawa$^{1}$, Takeshi Oishi$^{1}$ and Katsushi Ikeuchi$^{2}$% <-this % stops a space
	\thanks{$^{1}$ Ryoichi Ishikawa and Takeshi Oishi are with Institute of Industrial Science, The University of Tokyo,  Japan
		{\tt\small \{ishikawa, oishi\}@cvl.iis.u-tokyo.ac.jp}}%
	\thanks{$^{2}$Katsushi Ikeuchi is with Microsoft, USA,
		{\tt\small katsuike@microsoft.com}}%
}
\begin{document}

\maketitle
\thispagestyle{empty}
\pagestyle{empty}

%========================================================================================================
\begin{abstract}
In this paper, we propose a method of targetless and automatic Camera-LiDAR calibration. Our approach is an extension of hand-eye calibration framework to 2D-3D calibration. By using the sensor fusion odometry method, the scaled camera motions are calculated with high accuracy. In addition to this, we clarify the suitable motion for this calibration method.

The proposed method only requires
the three-dimensional point cloud and the camera image and does not need other information such as reflectance of LiDAR and to give initial extrinsic parameter.
In the experiments, we demonstrate our method using several sensor configurations in indoor and outdoor scenes to verify the effectiveness. The accuracy of our method achieves more than other comparable state-of-the-art methods.
\end{abstract}
%========================================================================================================
\section{Introduction}

\begin{figure}[t]
	\begin{center}
		\includegraphics[width=.9\linewidth]{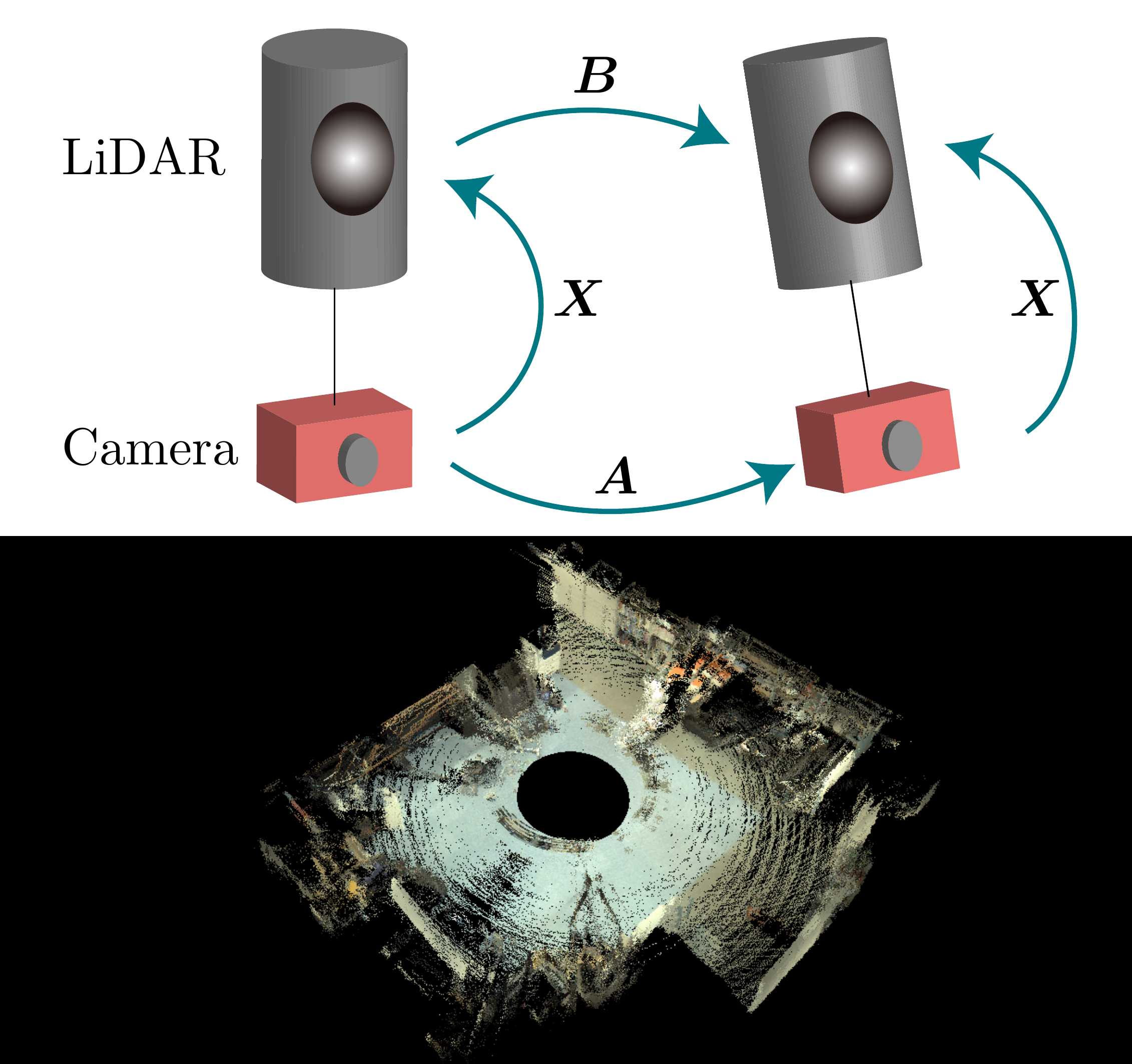}
		\caption{Top: Motion-based 2D-3D Calibration. In our method, the LiDAR motion is estimated by ICP algorithm and Camera motion is initially estimated by using feature point matching and then estimated with scale by sensor fusion system. Bottom: Colored scan by HDL-64E. Texture is taken by Ladybug 3 and our calibration result is used for texturing.}
		\label{fig:intro}
	\end{center}
\end{figure}

Sensor fusion has been widely studied in the field of robotics and computer vision. Compared to a single sensor system, higher level tasks can be performed by a fusion system combining multiple sensors. This type of system can be directly applied to three-dimensional environmental scanning. For example, by combining cameras with LiDAR, it is possible to perform color mapping on range images (Fig.~\ref{fig:intro}) or estimate accurate sensor motion for mobile sensing systems \cite{Inso_laserline,Zheng_balloon,ishikawa20163d,zhang2017real}. 

In a 2D-3D sensor fusion system composed of a camera and LiDAR, an extrinsic calibration of the sensors is required. There are methods that can obtain extrinsic parameter by using target cues or manually associating 2D points on the image with 3D points on the point cloud. However, manually establishing correspondences for accurate calibration is laborious because it requires multiple matches. Moreover, even though a lot of correspondence can be created in this way, the accuracy of calibration is still insufficient. Automated methods such as \cite{zhang2004extrinsic,fremont2008extrinsic} use targets that can be detected on both 2D images and 3D point clouds. However, since it is necessary to prepare targets detectable by both the camera and LiDAR, it is impractical and undesirable for on-site calibration. Recently, automatic 2D-3D calibration methods that do not require targets have been proposed. However, since the information obtained from each sensor is multi-modal, the calibration result depends on the modality between the sensors.

In this paper, we propose an automatic and targetless calibration method between a fixed camera and LiDAR. As shown in Fig.~\ref{fig:intro}, the proposed method is based on hand-eye calibration. In our method, the motions of the sensors are estimated respectively and calibration is performed using the estimated motions. Each sensor motion is calculated in the same modal and extrinsic parameter is derived numerically from each sensor motion. In the conventional motion-based 2D-3D calibration, the motion of the camera is obtained from only 2D images \cite{taylor2016motion}. However, the motion can only be estimated up to scale using only camera images. The precision of the extrinsic parameter is greatly affected by the motion error in the hand-eye calibration. Although the scale itself can be calculated simultaneously with extrinsic parameter from multiple motions, hand-eye calibration with scaleless motion deteriorates the accuracy of calibration.

On the other hand, in the sensor fusion odometry using the LiDAR and the camera, the motion of the camera can be accurately estimated with scale if the extrinsic parameter between the sensors are known \cite{Inso_laserline,Zheng_balloon,ishikawa20163d,zhang2017real}. In our method, we adopt the idea of camera motion estimation using sensor fusion odometry. First, an initial extrinsic parameter is obtained from scaleless camera motions and scaled LiDAR motions. Next, the camera motions are recalculated with scale using the initial extrinsic parameter and the point cloud from the LiDAR. Then the extrinsic parameter is calculated again using the motions. This recalculation of camera motions and recalculation of the extrinsic parameter are repeated until the estimation converges.

Our proposed method requires that the camera and the LiDAR have overlap in their measurement ranges and that the LiDAR's measurement range is 2D to align scans for the LiDAR motion estimation. The contributions of this paper are shown below.
\begin{itemize}
	\item As far as we know, this method is the first approach that incorporates camera motion estimation through sensor fusion into 2D-3D calibration.
	\item We study the optimal sensor motion which this calibration method works effectively.
	\item The input is only the RGB image from a camera and the three-dimensional point cloud from a LiDAR and does not need any other information such as the reflectance of a LiDAR or the initial value of the extrinsic parameter. The estimation result of the extrinsic parameter is more accurate than other methods with a small number of motion.
\end{itemize}
%========================================================================================================
\section{Related Works}
\label{sec:related}
Our work is related to the targetless and automatic 2D-3D calibration and hand-eye calibration.

\subsection{Target-less multi-modal calibration}
Targetless and automatic 2D-3D calibration methods generally use the common information existing in both image and point cloud. For example, portions that appear as discontinuous 3D shapes are highly likely to appear as edges on an RGB image. Methods that align this three-dimensional discontinuous portion with the 2D edge has been proposed \cite{levinson2013automatic,cui2016line}.
Meanwhile, a multi-modal alignment method using Mutual Information (MI) was proposed by Viola et al. \cite{viola}, and it has been developed mainly in the field of medical imaging. 2D-3D calibration methods using MI for evaluating the commonality between LiDAR and camera have also been proposed in recent years. As indicators evaluated through MI, reflectance - gray scale intensity \cite{pandey2015automatic}, surface normal - gray scale intensity \cite{taylor2012mutual}, and multiple evaluation indicators including discontinuities of LiDAR data and edge strength in images \cite{irie2016target} etc. are used. Taylor and Nieto also proposed gradient based metric in \cite{taylor2014multi}.

While these methods align 3D point cloud to 2D image using 3D to 2D projections, some texturing methods using images taken from multiple places with a camera to construct a stereo, reconstructing 3d structure from the images and aligning it to 3D point cloud has also been proposed. In \cite{Banno_cviu}, a method of computing the extrinsic parameter between the LiDAR and the camera for the texturing a dense 3D scan is proposed. The extrinsic calibration was done by aligning dense three-dimensional range data and sparse three-dimensional data reconstructed from two-dimensional stereo images.

\subsection{Hand-eye Calibration}
The method of changing the position and orientation of the sensor and performing the calibration using the motions observed by each sensor is known as hand-eye calibration.
Let ${\bf A}$ and ${\bf B}$ are the changes in position and orientation observed by two fixed sensors respectively, and ${\bf X}$ be the unknown relative position and orientation between sensors. Then the expression ${\bf AX}={\bf XB}$ holds(Refer top of Fig.~\ref{fig:intro}).
By using this expression to solve ${\bf X}$, extrinsic parameter between sensors can be obtained \cite{shiu1989calibration,fassi2005hand}.
Furthermore, due to the influence of noise on the sensor, Kalman Filter was used in calibration methods for estimating the bias of the sensor simultaneously \cite{hol2010modeling,kelly2011visual}.

In \cite{heng2013camodocal}, a method to calibrate the transitions between four cameras mounted on the car using visual odometry is proposed. In \cite{taylor2016motion}, Taylor and Nieto propose a method to obtain the motion of a sensor in 2D-3D calibration and estimate the extrinsic parameter and the time offset of the motion between the sensors. In their method, a highly accurate result is obtained by combining a multi-modal alignment method. However, since the method uses the scaleless position transition estimated from the camera images, it is difficult to obtain an accurate camera motion. In particular, it is difficult to accurately obtain the translation parameters with a small number of motions.
%========================================================================================================
\section{Methodology}

\begin{figure*}[t]
	\begin{center}
		\includegraphics[width=.9\linewidth]{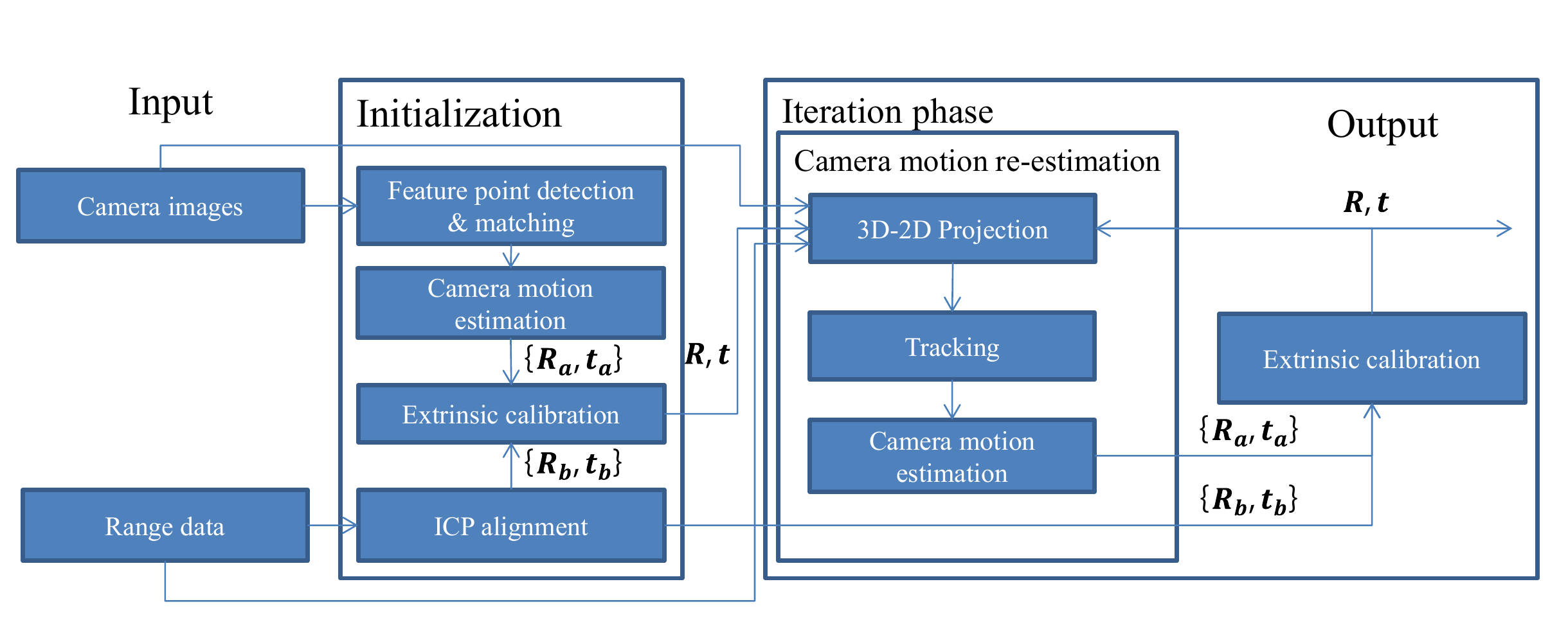}
		\caption{Overview}
		\label{fig:overview}
	\end{center}
\end{figure*}

Figure~\ref{fig:overview} shows the overview of our method. This method is roughly divided into two steps. In the initialization phase, we estimate the extrinsic parameter from the LiDAR motions using ICP alignment and camera motions using feature point matching. Then we alternatingly iterate estimating the extrinsic parameter and the camera motions through sensor fusion odometry. 

%Once the extrinsic parameter is determined, 2D-3D correspondence at the measurement time $t_1$ can be created by projecting the range data to the image using extrinsic parameters.

%Next, 2D-2D correspondence can be obtained by tracking the points on the image of $t_1$ where the three-dimensional points are projected to the camera frame of another time $t_2$.

%Finally, by integrating these 2D-3D correspondences and 2D-2D correspondences, 3D-2D correspondence can be made between the three-dimensional point of the range data at the time $t_1$ and the point on the image at the time $t_2$. By using correspondence, it is possible to calculate the relative position and orientation of the cameras at time $t_1$ and time $t_2$.

\subsection{Initial calibration parameter estimation}
\subsubsection{Sensor motion estimation}
First, we explain about the motion estimation of each sensor in the initial extrinsic parameter estimation.

{\bf LiDAR}

For the estimation of the LiDAR motion, we use a high-speed registration method through ICP algorithm which searches correspondence points in gaze directions \cite{alignment}. We create meshes on point clouds in advance using a sequence of points and project these points onto two dimension and use Voronoi splitting. When aligning scans, initially, the threshold value of the distance between corresponding points is set to be large, and the outlier correspondences are eliminated while gradually decreasing the threshold value.

{\bf Camera}

For the initial motion estimation of the camera, a method using standard feature point matching is used. First, we extract feature points from two images using AKAZE algorithm \cite{alcantarilla2011fast}, calculate descriptors, and make matchings. From the matchings, we calculate the initial relative position and pose between camera frames using 5 point algorithm \cite{Nister_2006} and RANSAC \cite{fischler1981random}. After obtaining the initial relative position and orientation, optimization is performed by minimizing the projection error using an angle error metric with the epipolar plane was used \cite{pagani2011structure}.

\subsubsection{Initial extrinsic parameter calibration from sensor motion}
In order to obtain the relative position and orientation between the two sensors from the initially estimated motions, we use a method that extends the normal hand-eye calibration to include the estimation of the camera motion's scale.
Let the $i$ th position and pose changes measured by the camera and the LiDAR be $ 4 \times 4 $ matrix $ {\bf A^i} $, ${\bf B^i}$ respectively and extrinsic parameter between two sensors be $ 4 \times 4 $ matrix ${\bf X}$. ${\bf A^iX}={\bf X^iB}$ can be established and the following two equations hold by decomposing it \cite{shiu1989calibration},
\begin{eqnarray}
\label{eq:axis}
{\bf R}_a^{i}{\bf R}&=&{\bf R} {\bf R}_b^{i}\\
\label{eq:trans}
{\bf R}_{a}^{i}{\bf t}+{\bf t}_a^{i}&=&{\bf R}{\bf t}_b^{i}+{\bf t},
\end{eqnarray}
where ${\bf R}_{a}^{i}$ and ${\bf R}_{b}^{i}$ is a $3 \times 3$ rotation matrix of ${\bf A}^{i}$ and ${\bf B}^{i}$, ${\bf t}_a^{i}$ and ${\bf t}_b^{i}$ represents the $3 \times 1$ vector of translational component of ${\bf A}^{i}$, ${\bf B}^{i}$. Let ${\bf k}_a^{i}$ and ${\bf k}_b^{i}$ be rotational axis of rotation matrix ${\bf R}_{a}^{i}$ and ${\bf R}_{b}^{i}$. When Eq.~\ref{eq:axis} holds, following equation holds,
\begin{eqnarray}
\label{eq:axis2}
{\bf k}_a^{i}={\bf R} {\bf k}_b^{i}.
\end{eqnarray}
Since the absolute scale of translational movement between camera frames can not be calculated, Eq.~\ref{eq:trans} is written as following using the scale factor $s^{i}$,
\begin{eqnarray}
\label{eq:trans2}
{\bf R}_{a}^{i}{\bf t}+s^{i} {\bf t}_a^{i}={\bf R}{\bf t}_b^{i}+{\bf t}.
\end{eqnarray}

${\bf R}$ is linearly solved by using SVD from series of ${\bf k}_a^{i}, {\bf k}_b^{i}$. However, to solve the rotation, more than two position and pose transitions are required and rotations in different directions must be included in the series of transition. In nonlinear optimization, ${\bf R}$ is optimized by minimizing the following cost function derived from Eq.~\ref{eq:axis},
\begin{eqnarray}
\label{eq:rotopt}
{\bf R}=\argmin_{R}\sum_{i}\left|{\bf R}_a^{i}{\bf R}-{\bf R} {\bf R}_b^{i}\right|
\end{eqnarray}
After optimizing ${\bf R}$, ${\bf t}$ and $s^{i}$ is obtained by constructing a simultaneous equation and solving it linearly using the least squares method.

\subsection{Iteration of Camera motion estimation and Calibration from sensor motion}
Initial extrinsic parameter can be obtained using estimated LiDAR motions and initial camera motions. However, scale information of camera motion cannot be obtained from camera images alone. The rotation component of the extrinsic parameter is independent of the scale information and can be accurately estimated in the initial parameter estimation phase. On the other hand, the translation of the extrinsic parameter can be calculated from the difference between the movement amount of the camera and the LiDAR when rotating sensors as indicated in Eq.~\ref{eq:trans}. Therefore, since the precision of the translational component of the extrinsic parameter is deeply related to the accuracy of camera motion estimation, it is difficult to accurately estimate the extrinsic parameter from the scaleless camera motion.

On the other hand, the motion estimation using the sensor fusion system can solve the scaled motion with high accuracy \cite{Inso_laserline, Zheng_balloon}. Once the extrinsic parameter is estimated, we can estimate the camera motion ${\bf A^{i}}$ with scale by using the given extrinsic parameter ${\bf X}$ and the range data scanned by the LiDAR. After motion estimation, the extrinsic parameter $ {\bf X} $ is re-estimated using the series of $ {\bf A^{i}} and {\bf B^{i}} $. Since there is no need to estimate the translation component of $ {\bf A^{i}} $ at the same time, it is possible to estimate the translation component of $ {\bf X} $ more accurately than the initial estimation.
Estimating the camera motion $ {\bf A^{i}} $ again using the re-estimated $ {\bf X} $ and the range data increases the accuracy of $ {\bf A^{i}}$. The extrinsic parameter is then optimized by repeating the estimation of the camera motion and the estimation of the extrinsic parameter alternately until convergence.

%In the alternative repetition of re-estimation of camera motion and re-estimation of external parameters, it is inevitably to estimated each parameter with an error in the mutual parameter. Therefore, after giving the detailed explanation of re-estimation of camera motion and re-estimation of external parameters in this section, we explain the influence of parameter error on the mutual parameter estimation in the next section. Then we describe the condition of the sensor motion which is good for parameter convergence.
\subsubsection{Camera motion estimation with range data}

\begin{figure}[t]
	\begin{center}
		\includegraphics[width=.8\linewidth]{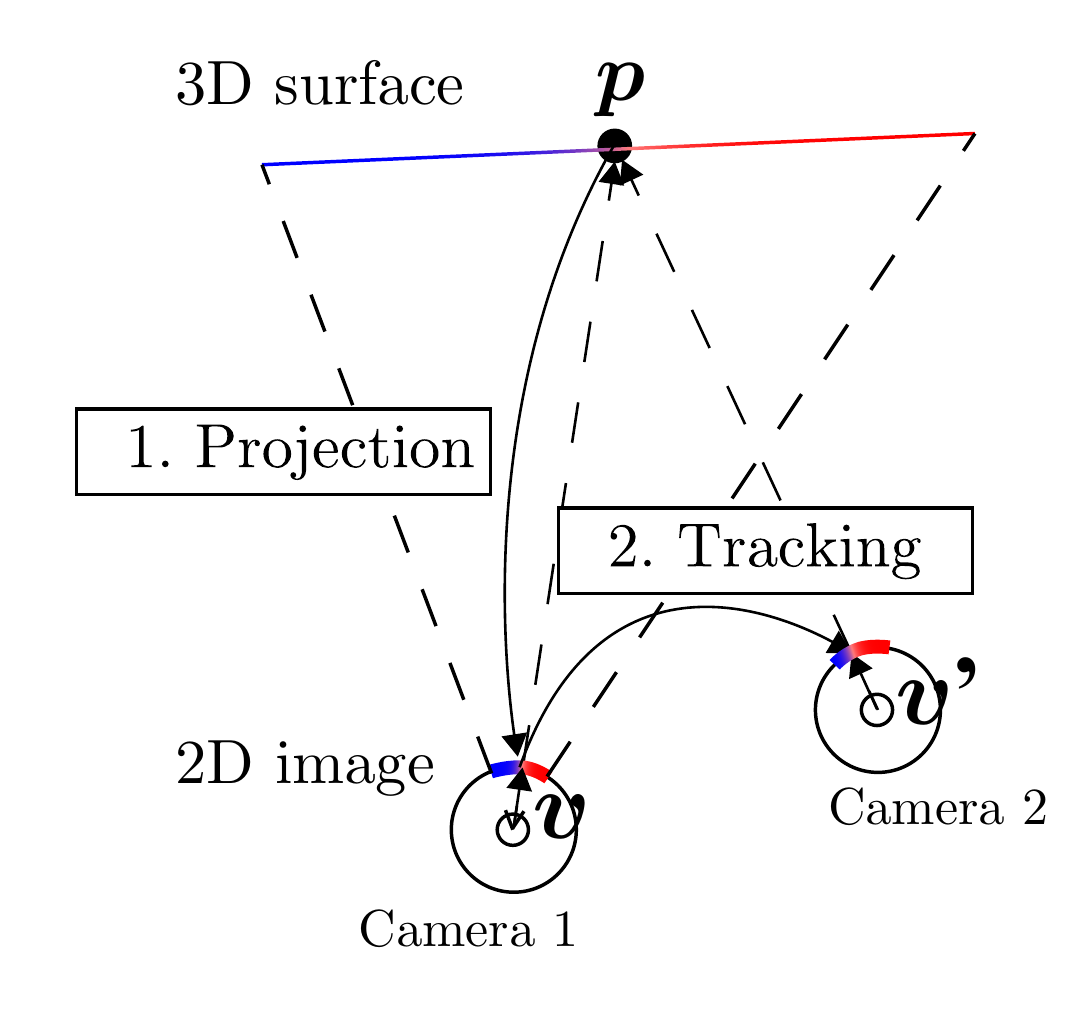}
		\caption{Schematic diagram of how to obtain 2D-3D correspondences}
		\label{fig:proj2d3d}
	\end{center}
\end{figure} 

Figure~\ref{fig:proj2d3d} shows that the schematic diagram of constructing 2D-3D correspondence. Inputs are point cloud in the world coordinates, two camera images taken at the position of camera 1, 2, and extrinsic parameter to localize camera 1 into the world coordinates. First, a certain point ${\bf p}$ in the point cloud onto the image in Camera 1 using projection function $Proj({\bf p}_c)$ which project 3D points ${\bf p}_c$ in camera coordinates onto camera image and extrinsic parameter by following equation, 
\begin{eqnarray}
{\bf v}=Proj({\bf Rp}+{\bf t}),
\end{eqnarray}
where ${\bf v}$ is the vector heading from the center of camera 1 to the corresponding pixel. Then we track the pixel on which ${\bf p}$ is projected from camera image 1 to image 2 using KLT tracker \cite{KLT_1981}. Let ${\bf v}'$ be the vector heading from the center of camera 2 to the tracked pixel. Now the point ${\bf p}$ and the vector $ {\bf v}'$ of 2D-3D correspondence is constructed.

After constructing 2D-3D correspondences, it is possible to optimize the relative position and orientation of the camera 1 and the camera 2 by minimizing the projection error. Let $ ({\bf v}'_j, {\bf p}_j)$ be the $j$ th 2D-3D correspondence in $i$ th motion, the position and pose transition between cameras $ {\bf R}_a^i, {\bf t }_a^i $ can be optimized by minimizing the following angle metric cost function.
\begin{eqnarray}
{\bf R}_a^i, {\bf t}_a^i=\argmin_{R_a,t_a} \sum_{j}\left|v'_j\times Proj({\bf R_a}({\bf R}{\bf p}_j+{\bf t})+{\bf t_a})\right|
\end{eqnarray}
For the initial values of ${\bf R}_a$ and ${\bf t}_a$, generalized perspective 3 point algorithm \cite{kneip2011novel} is used in the first iteration. After the second iteration, the estimation result in the previous iteration is used.

\subsubsection{Parameter Calibration}
Once the position and pose transition of the camera is recalculated, the extrinsic parameter is optimized again using the motion of the camera and the LiDAR. In each iteration, $ {\bf R} $ and ${\bf t} $ is solved linearly and non-linearly. In non-linear optimization,  ${\bf R}$ is optimized by Eq.~\ref{eq:rotopt} and ${\bf t} $ is optimized by following,
\begin{eqnarray}
{\bf t}=\argmin_{t}\sum_{i}{\left|({\bf R}_a^i{\bf t}+{\bf t}_a^i)-({\bf R}{\bf t}_b^i+{\bf t})\right|}
\end{eqnarray}

%===========================================================================================================

\section{Optimal motion for 2D-3D calibration}
We consider the motion suitable for the calibration taking into consideration the influence each other's error has on the mutual parameter estimation. During the alternating estimation of the extrinsic parameter between the sensors and the motion of the camera, it is inevitable to estimate the position and pose of each other with the error. Since motion estimation and extrinsic parameter estimation are also dependent on the measured environment and the number of motions, it is difficult to obtain precise convergence conditions. However, it is possible to consider the motion that is likely to converge the estimation. 
\subsection{Camera motion estimation}

\begin{figure}[t]
	\begin{center}
		\includegraphics[width=\linewidth]{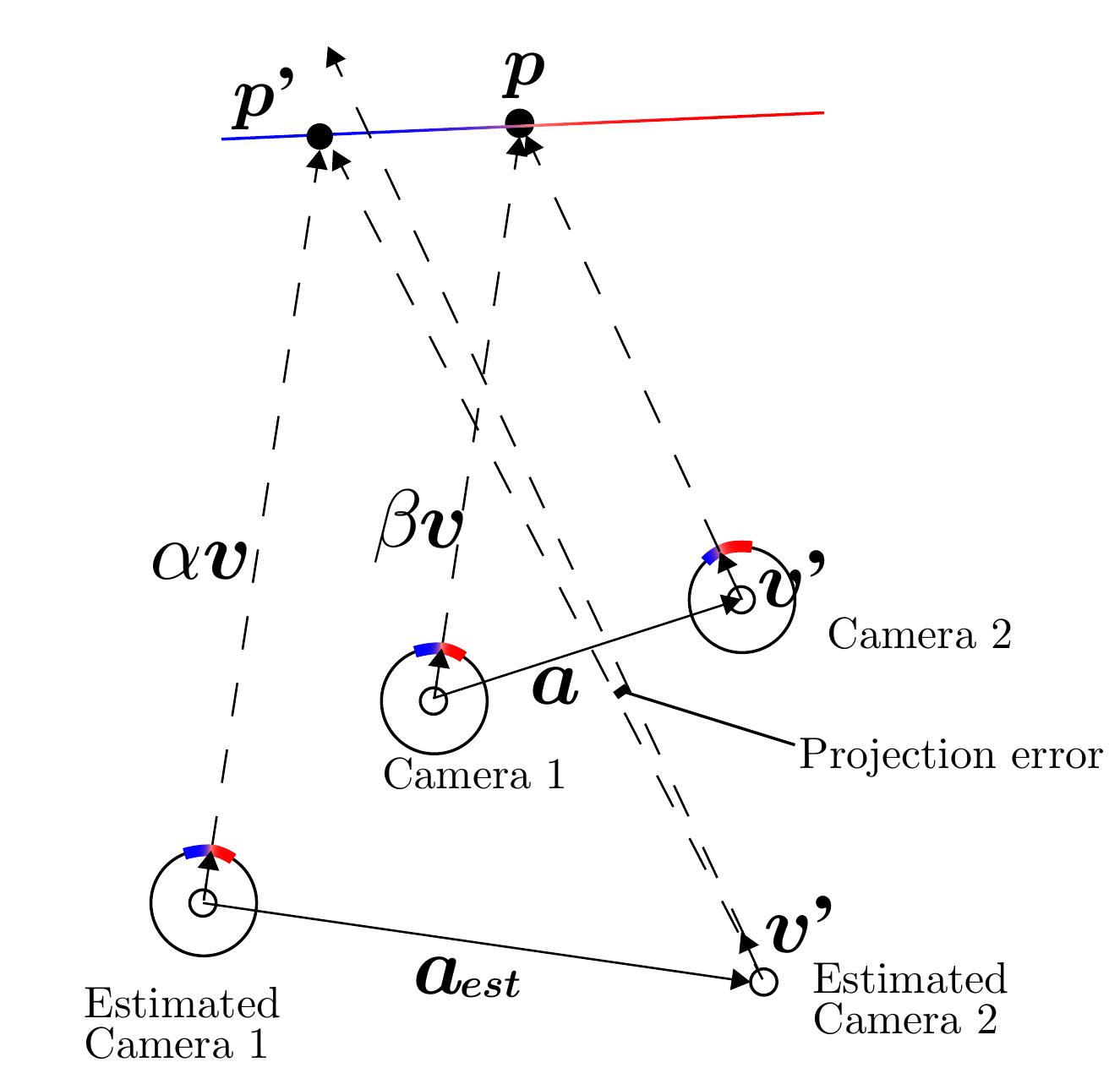}
		\caption{Schematic diagram when there is an error in the localized position of the camera 1}
		\label{fig:camLocalize}
	\end{center}
\end{figure}

First, we consider the influence of the extrinsic parameter error on the localization of the camera and the conditions under which the estimation is successful. We suppose the case where there is an error in the extrinsic parameter given in the section as shown in Fig.~\ref{fig:camLocalize}. Let "Estimated Camera 1" be the estimated position of camera 1 with respect the actual camera position (Camera 1, 2 in Fig.~\ref{fig:camLocalize}). The rotation error between Camera 1 and Estimated Camera 1 is considered to be negligibly small.

Next, consider the operation of creating 2D-3D correspondence. In the projection step, a certain point ${\bf p} '$ on the point cloud is projected onto the Estimated camera 1. However, a difference is caused between the projected pixel and the 3D point due to the error of the extrinsic parameter. Then the pixel on which the point $ {\bf p} '$ is projected is tracked on the Camera 2 image. Let $ {\bf v} $ be the vector heading from Estimated camera 1 to the point $ {\bf p}' $. Point $ {\bf p}$ is actually corresponds of the vector $ {\bf v} $. Ignoring the error of pixel tracking from Camera 1 to Camera 2, the direction vector from Camera 2 to the pixel corresponding to ${\bf v}$ is $ {\bf v}'$. On the computer, the three-dimensional point ${\bf p}'$ and $ {\bf v}' $ correspond to each other.

For the projection error, assuming that there is no rotation error in estimating the position and orientation of the camera 2, the projection error when the estimated position is $ {\bf a}_{est} $ is
\begin{eqnarray}
\label{eq:projerr}
e({\bf a}_{est})= \left|{\bf v}' \times \frac{\alpha {\bf v}-{\bf a}_{est}}{|\alpha {\bf v}-{\bf a}_{est}|} \right|.  
\end{eqnarray}
From Fig.~\ref {fig:camLocalize}, let ${\bf a}$ be the vector directing from Camera 1 to Camera 2, ${\bf v}'$ is expressed as following,
\begin{eqnarray}
\label{eq:vd}
{\bf v}'=\frac{\beta {\bf v}-{\bf a}}{|\beta {\bf v}-{\bf a}|}.
\end{eqnarray}
Substitute Eq.~\ref{eq:vd} for Eq.~\ref{eq:projerr},
\begin{eqnarray}
\label{eq:projerr2}
e({\bf a}_{est})= \left| \frac{\beta {\bf v}-{\bf a}}{|\beta {\bf v}-{\bf a}|} \times \frac{\alpha {\bf v}-{\bf a}_{est}}{|\alpha {\bf v}-{\bf a}_{est}|} \right|.
\end{eqnarray}
When optimizing ${\bf a}_{est}$, we compute the projection error for ${\bf v}$ in all directions and ${\bf a}_{est}$ approach to the point where the sum of the projection error is minimized.

Now, in order to estimate the extrinsic parameter accurately, ideal camera motion should be estimated as real camera actually moves.
In other words, ideally estimated motion becomes $ {\bf a}_{est} \to {\bf a} $. In order for $ {\bf a}_{est}$ to approach $ {\bf a} $, the projection error $e({\bf a})$ when Estimated Camera 2 is located away from Estimated Camera 1 by ${\bf a}$ becomes small. $e({\bf a})$ is represented by the following equation:
\begin{eqnarray}
\label{eq:projerr3}
e({\bf a})&=&\left| \frac{(\alpha-\beta) ({\bf a}\times{\bf v})}{|\beta {\bf v}-{\bf a}||\alpha {\bf v}-{\bf a}|} \right|.
\end{eqnarray}

From Eq.~\ref{eq:projerr3}, the followings can be said.
\begin{itemize}
	\item The smaller the ${\bf a}$, the smaller the projection error even if the extrinsic parameter contains error. That is, the smaller the moving distance of the camera is, the more accurate the estimation becomes. 
	In other words, when ${\bf a}$ is small, the existence probability of Estimated Camera 2 appears to the periphery of the ideal place acutely. Therefore in the subsequent extrinsic parameter estimation, the existence probability of Estimated Camera 1 also appears sharply around the true value.
	\item The smaller the value of $ (\alpha - \beta)$, the smaller the projection error. The cases the difference of $ (\alpha - \beta) $ comes out are, for example, when there is a large step under environments or when the incident angle from the camera is shallow. Therefore it is preferable that the calibration environment is, for example, surrounded by smooth walls.
\end{itemize}

\subsection{Extrinsic parameter estimation}
Next, we consider the influence of the error in the estimated motions on extrinsic parameter calibration. Ignoring the rotation error for the simplification, consider the case where there is the error in the translation of the camera motion and the translation of the extrinsic parameter for Eq.~\ref{eq:trans}. Let $ {\bf e}_a$ and ${\bf e}$ be the error with respect to ${\bf t}_a$ and $ {\bf t} $,
\begin{eqnarray}
\label{eq:trans_err}
{\bf R}_{a}({\bf t}+{\bf e})+{\bf t}_a+{\bf e}_a={\bf R}{\bf t}_b+{\bf t}+{\bf e}.
\end{eqnarray}
Taking the difference between Eq.~\ref{eq:trans_err} and Eq.~\ref{eq:trans},
\begin{eqnarray}
\label{eq:diff}
{\bf e}_a=({\bf I}-{\bf R}_{a}){\bf e}.
\end{eqnarray}
In the case seeing Eq.~\ref{eq:diff} as a single unit, when the rotation amount of the ${\bf R}_{a} $ is small, the translation errors become $ |{\bf e}_a|<| {\bf e} | $. This indicates that the error of the camera motion propagates to the extrinsic parameter in the diverging direction to the error.

If the amount of error propagation in estimating the extrinsic parameter from the camera motions does not exceed the amount of the error reduction in the camera motion estimation using the distance image, the accuracy of the relative position and orientation is improved by the proposed method. Therefore, in order to reduce the propagation amount of error in relative position and pose estimation, increasing the rotation amount of the camera motion is effective. It is also effective to sample a plurality of motions as much as possible for robust extrinsic parameter estimation. Regarding the amount of rotation of the camera motion, if the appearance of the image changes significantly, it might affect the accuracy of the motion estimation. Therefore this also needs to be taken into account. Although the proposed method can be applied to perspective camera, an omnidirectional camera has advantage because it is possible to secure a common field of view even when the camera rotates significantly.
%========================================================================================================
\section{Experimental results}
\begin{figure}[t]
	\begin{center}
		\includegraphics[width=\linewidth]{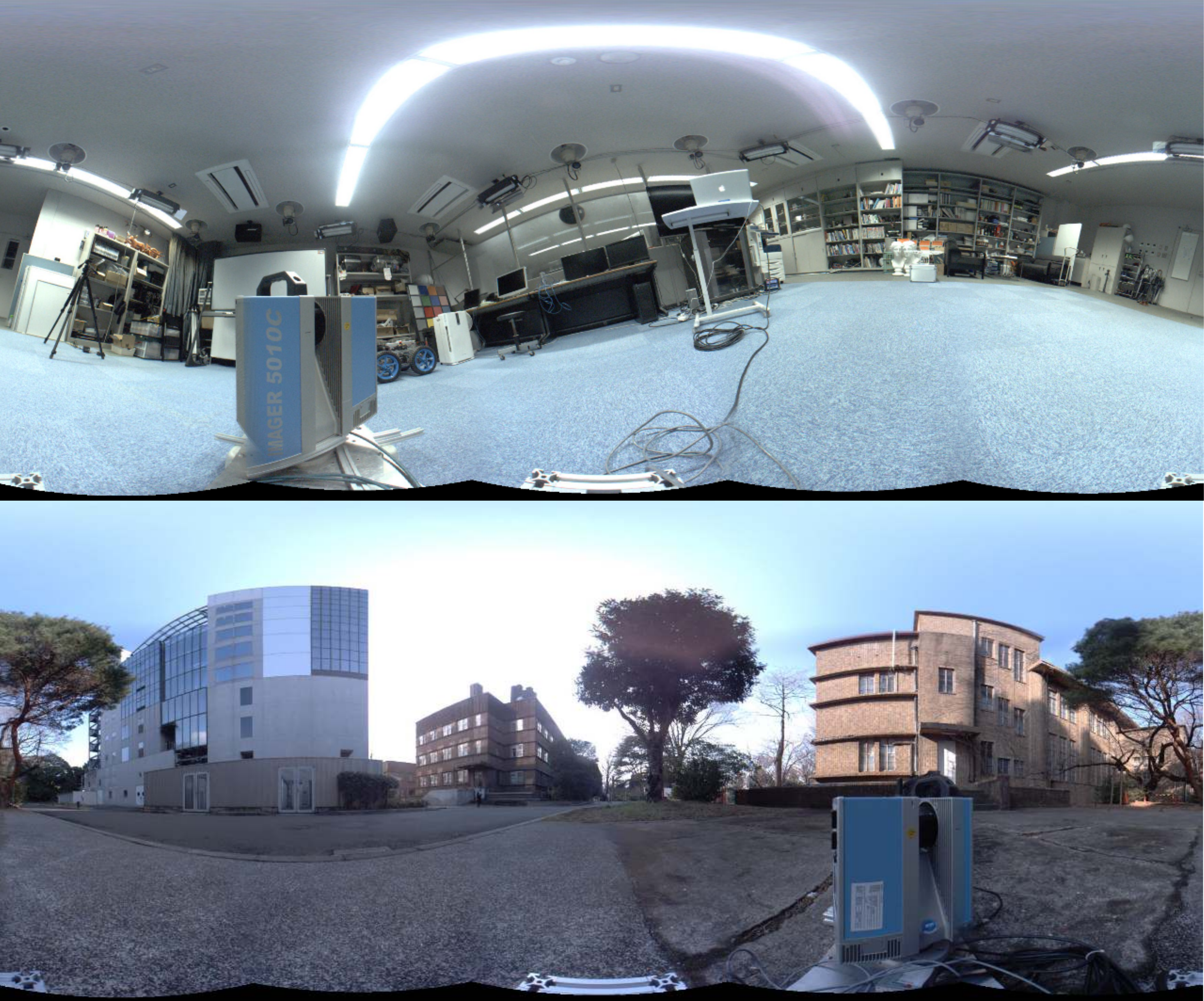}
		\caption{Indoor and outdoor calibration scene taken by Ladybug 3}
		\label{fig:scene}
	\end{center}
\end{figure}

\begin{figure}[t]
	\begin{center}
		\includegraphics[width=\linewidth]{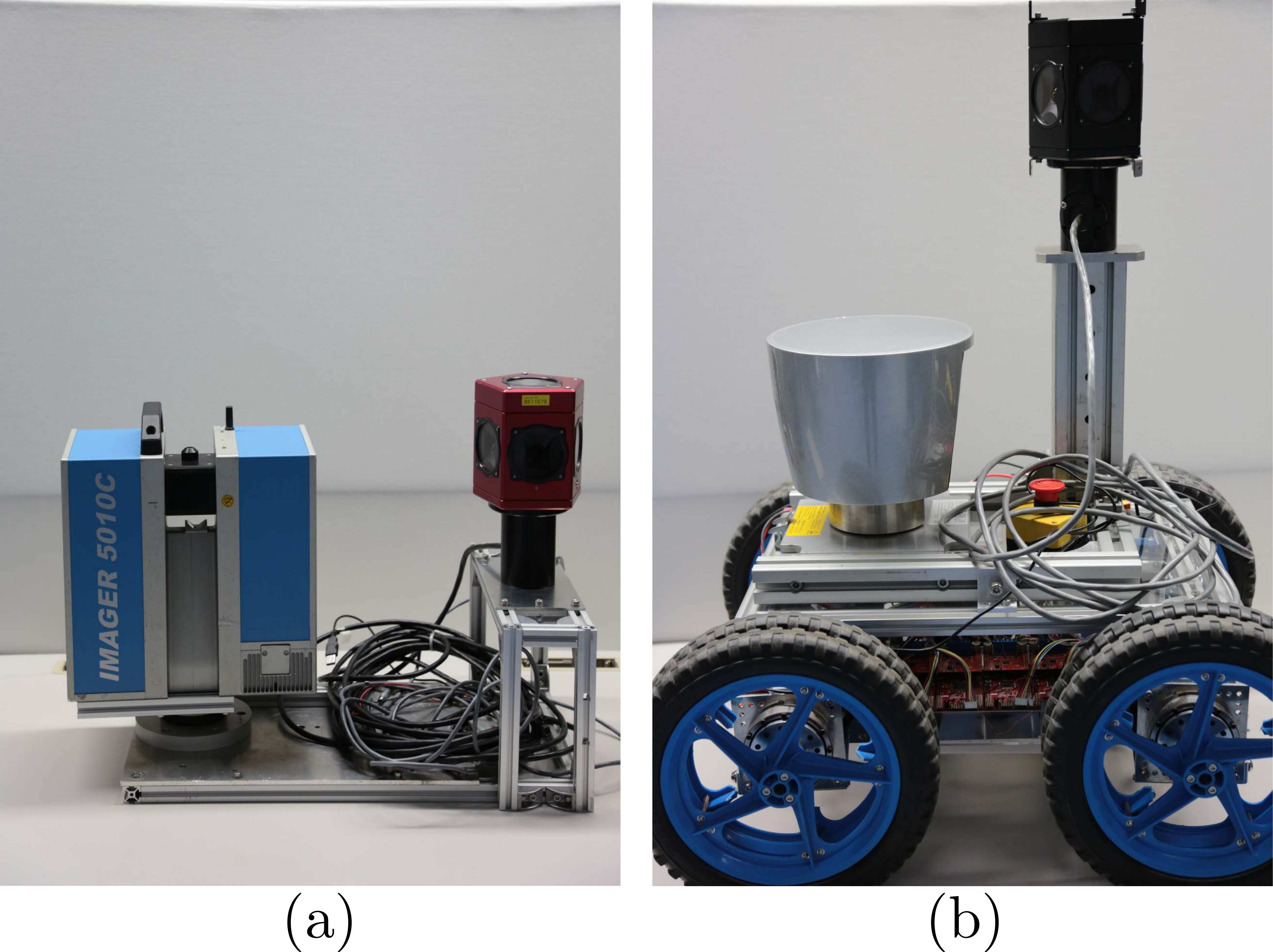}
		\caption{Sensor configuration. (a)Imager 5010C and Ladybug 3, (b)HDL-64E and Ladybug 3}
		\label{fig:sensorconfig}
	\end{center}
\end{figure}

In the experiments, we conduct calibrations in indoor and outdoor environments shown in Fig.~\ref{fig:scene} using panoramic LiDAR, multi-beam LiDAR, and omnidirectional camera. One of the compared methods is image-based calibration using the camera motions with no scale, which is used as the base in \cite{taylor2016motion}. In Fig.~\ref{fig:overview}, this is the initialization output and hereinafter referred to it as "{\it Scaleless}". In addition to {\it Scaleless}, we used calibration by {\it Manual} correspondence acquisition and calibration with alignment using {\it MI} \cite{pandey2015automatic} as the other compared methods.
\subsection{Evaluation with colored range data}
First, the results using the evaluation datasets are shown.
To measure the data set, two range sensors Focus S 150 by FARO Inc.\footnote{https://www.faro.com} and Imager 5010C by Zollar+Fl\"{o}hlich Inc.\footnote{http://www.zf-laser.com} are used. Three-dimensional panoramic point clouds are scanned with both range sensors. In the data measured by Focus S 150, a colored point cloud is obtained using a photo texture function of it. In the evaluation, inputs are a pseudo panorama rendered image obtained from colored point cloud scanned by Focus S 150 and a point cloud scanned by Imager 5010C. Ground truth is computed through registration of two point clouds scanned by the two sensors.

In the indoor scene dataset, motions are obtained by rotating sensors 5 times in the vertical direction and 5 times in the horizontal direction to measure the data.
\begin{figure}[t]
	\begin{center}
		\includegraphics[width=.8\linewidth]{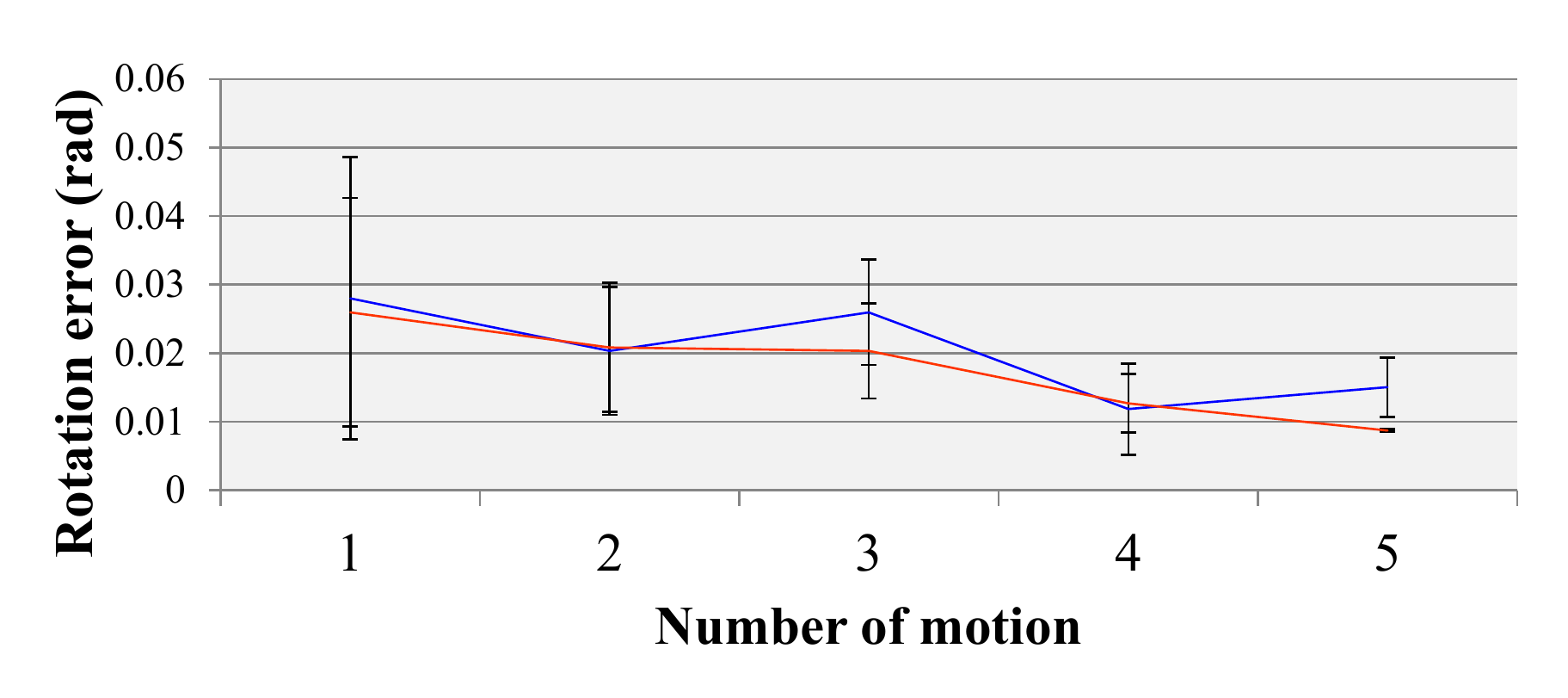}
		\caption{Transition graph of rotation error when changing the number of motion. Blue line: {\it Scaleless}, Red line: Ours.}
		\label{fig:graphr}
	\end{center}
\end{figure}
\begin{figure}[t]
	\begin{center}
		\includegraphics[width=.8\linewidth]{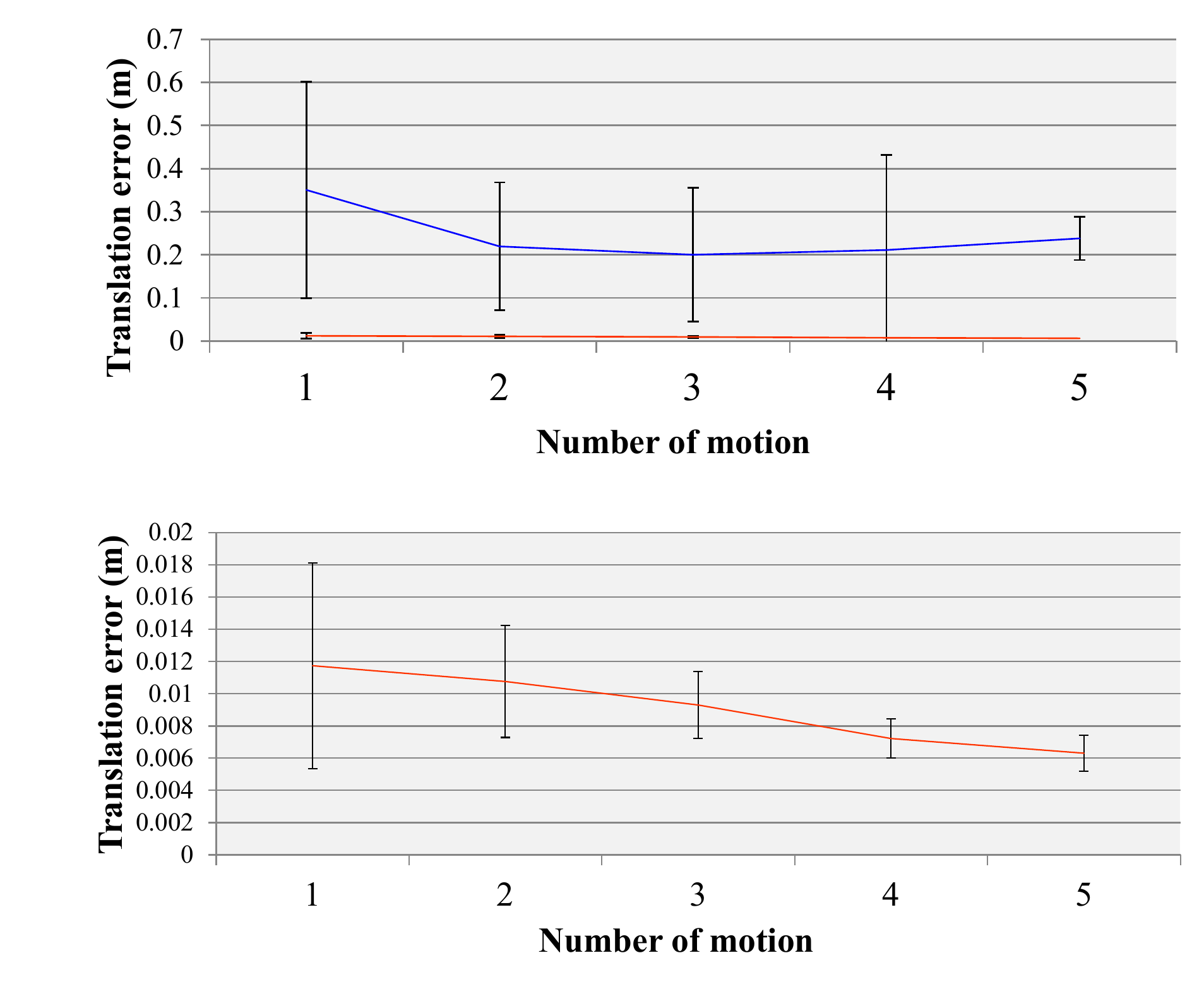}
		\caption{Transition graph of translation error when changing the number of motion. Blue line: {\it Scaleless}, Red line: Ours. Bottom shows the result of our method only.}
		\label{fig:grapht}
	\end{center}
\end{figure}
The result of calibration with changing the number of motion in the indoor scene is shown in Fig.~\ref{fig:graphr} and Fig.~\ref{fig:grapht}.
The horizontal axis of the graphs indicates the number of horizontal and vertical motions used for calibration. For example, when the number is one, it indicates that the calibration is performed using two motions in total with one horizontal and one vertical motion. 
Evaluation is performed by conducting calibration 10 times in each motion number sampling motions randomly. Figure~\ref{fig:graphr} and Fig.~\ref{fig:grapht} show the graphs plotting the average and standard deviation of the error of rotation and translation of extrinsic parameter. The blue line indicates the error of the {\it Scaleless} and the red line indicates the error of the extrinsic parameter estimated by ours. In terms of rotation, there is no great improvement in accuracy. However, for the translation error, the accuracy improves dramatically using the proposed method as shown in Fig.~\ref{fig:grapht}. It is also shown that the error is gradually decreasing by increasing the number of motions.

\begin{figure}[t]
	\begin{center}
		\includegraphics[width=\linewidth]{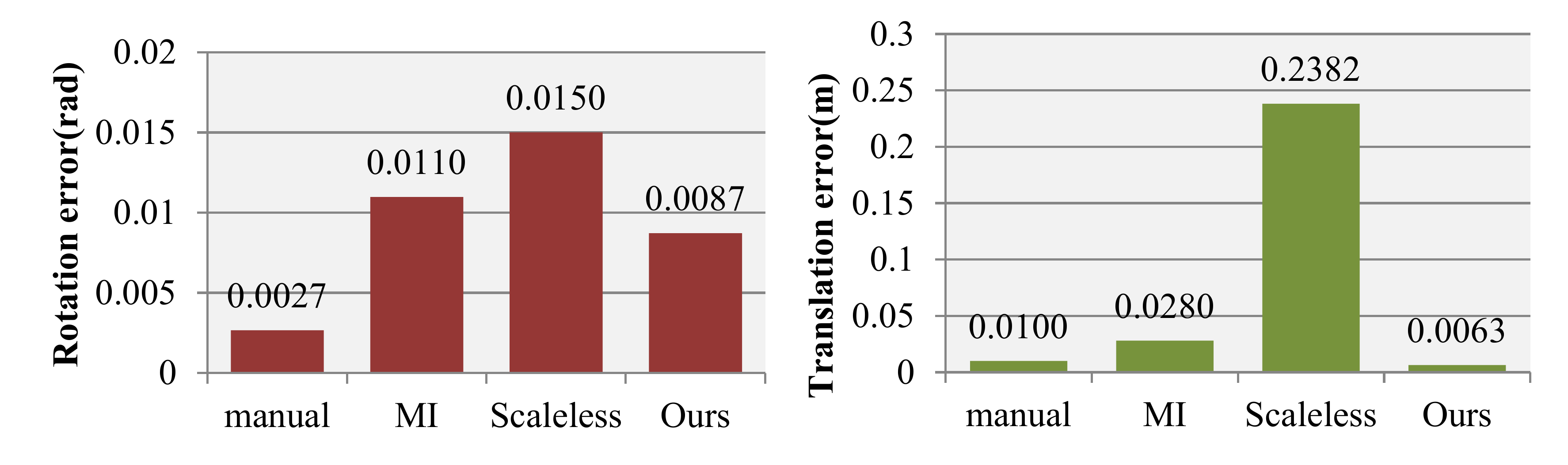}
		\caption{Error from the ground truth of the calibration result by each method in indoor scene}
		\label{fig:indoorcomp}
	\end{center}
\end{figure}

The results compared with other methods ({\it Manual}, {\it MI}) are shown in Fig.~\ref{fig:indoorcomp}. In the registration by maximizing MI, only 1 scan is used, and the initial point was shifted from the ground truth by a fixed distance ($0.1 m$) in the random direction only for the translation parameter. In the {\it Manual} calibration, calibration is carried out by acquiring 30 correspondings as much as possible from all directions.
From the Fig.~\ref{fig:indoorcomp}, the rotational error is less than 1 degree in any method. While, in  translational error, ours achieves the least error compared to other methods.

\begin{figure}[t]
	\begin{center}
		\includegraphics[width=\linewidth]{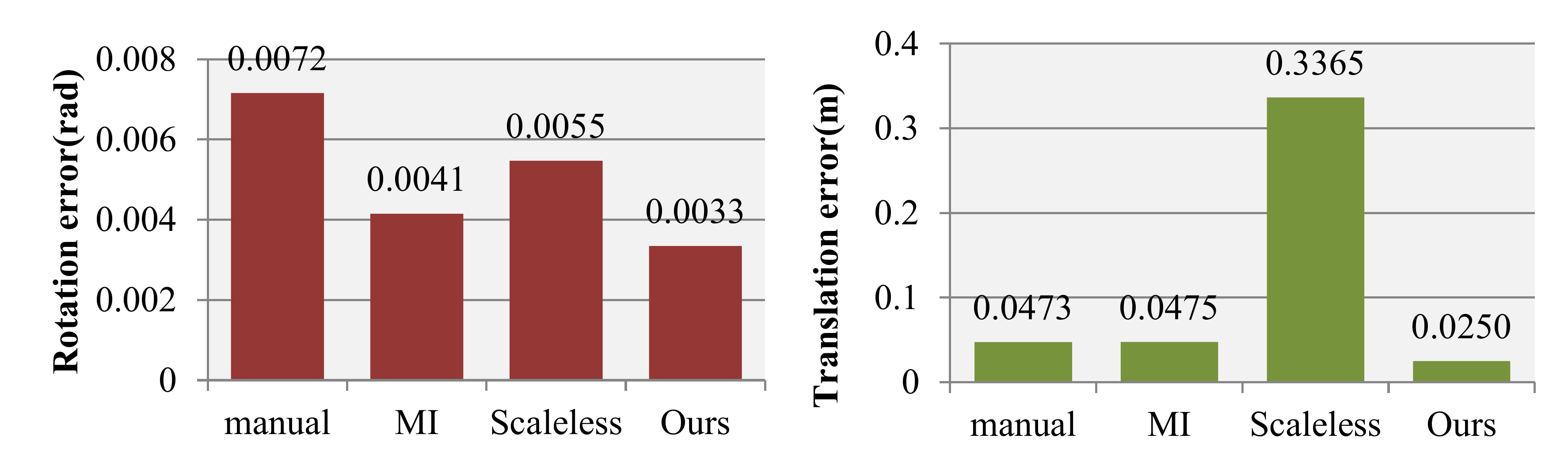}
		\caption{Error from the ground truth of the calibration result by each method in outdoor scene}
		\label{fig:outdoorcomp}
	\end{center}
\end{figure}

Evaluation results using the outdoor environment dataset are also shown in Fig.~\ref{fig:outdoorcomp}. Motions are obtained by rotating sensors by 3 times in the vertical direction and 3 times in the horizontal direction to measure the data. For the rotation, our method obtained the best result. However, in any method, the errors are less than 0.5 degrees and no significant difference is seen. On the other hand, ours has the best estimation result for the translation. Considering that the accuracy is less than 1 cm with the same number of motions in the indoor environment, we can say that the indoor scene is good for calibration.

%========================================================================================================
\subsection{Ladybug 3 and Imager 5010C}
\begin{figure}[t]
	\begin{center}
		\includegraphics[width=\linewidth]{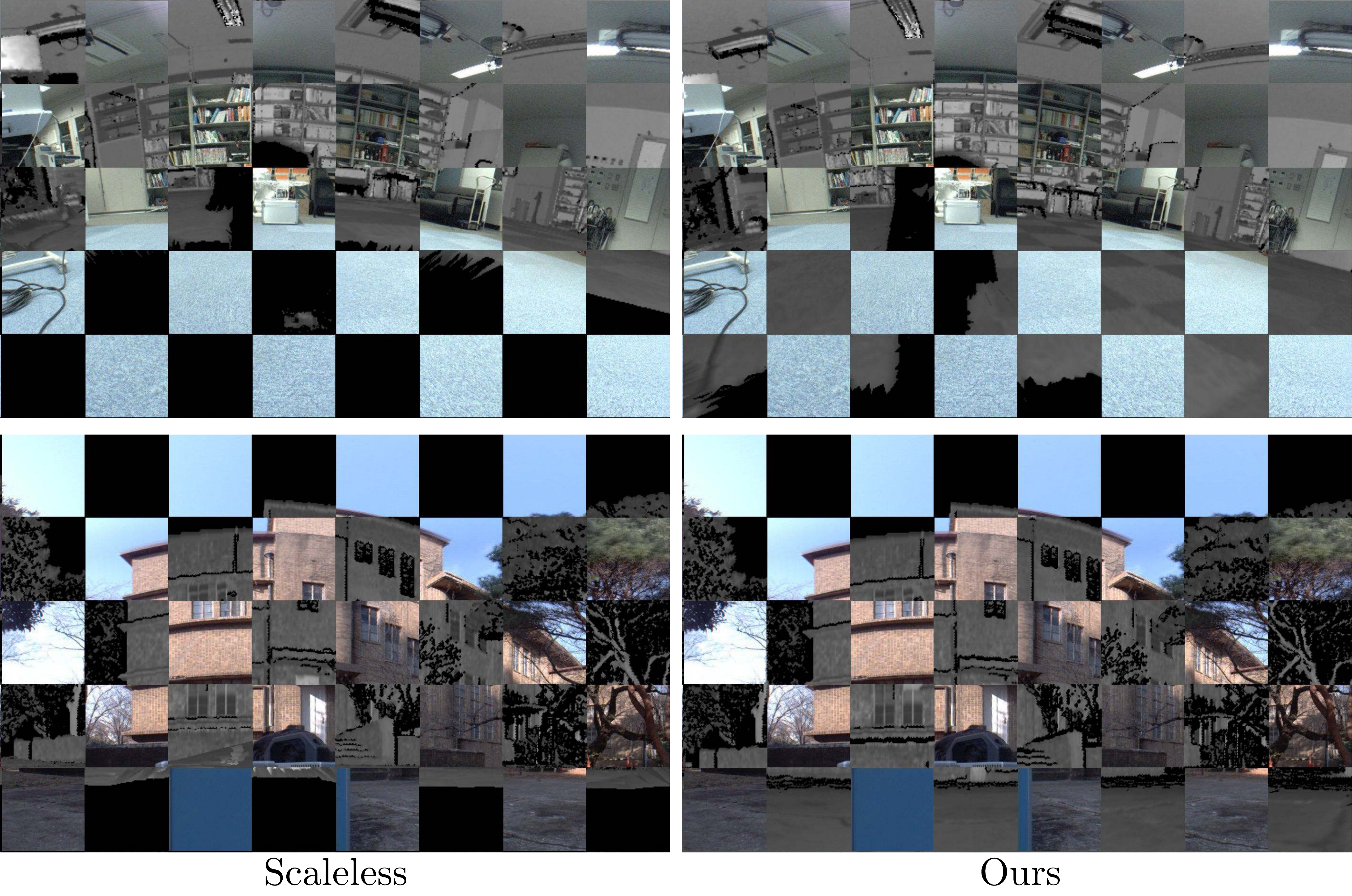}
		\caption{The pictures in which panorama images taken by Ladybug 3 and panorama rendered reflectance image scanned by Imager 5010C are alternately arranged like a checker. When extrinsic parameter is correct, consistency is established between the two images. We set stitching distance of panoramic image to $4 m$ in indoor scene and $7 m$ in outdoor scene.}
		\label{fig:zflb}
	\end{center}
\end{figure}
Next, we show the results of calibration using omnidirectional camera Ladybug 3 by FLIR Inc.\footnote{https://www.ptgrey.com/} and Imager 5010C. The appearance of the sensor configuration is shown in Fig.~\ref{fig:sensorconfig}~(a). 

For the evaluation, as shown in Fig.~\ref{fig:zflb}, the evaluation is performed by overlaying the image of Ladybug 3 and the reflectance image obtained by panorama rendering from the center of the estimated camera position in the point cloud. Images are displayed alternately in the checker pattern. We then confirm the consistency on the two images. From Fig.~\ref{fig:zflb}, {\it Scaleless} before optimization does not have consistency between RGB image and reflectance image, but the results of the proposed method have consistency between two images.

\subsection{Ladybug 3 and Velodyne HDL-64E}
We show the result of extrinsic calibration of HDL-64E by Velodyne Inc. \footnote{http://velodynelidar.com/} which is multi-beam LiDAR and Ladybug 3 in the indoor scene. The appearance of the sensor configuration is shown in Fig.~\ref{fig:sensorconfig}~(b). In the measurement of data, the rover loading the sensors is operated to generate the rotation motion in the vertical direction and the horizontal direction. For the rotation in the vertical direction, motions are generated by raising and lowering only the front wheel by about 4 cm step. When acquiring data, we stopped the rover and measured it with stop-and-scan. To obtain range data and camera image scanned at the same position, we visually check the timing when the LiDAR and the camera were stationary.

The ground truth of extrinsic parameter between HDL-64E and Ladybug 3 is indirectly obtained by using the point cloud measured by Imager 5010C under the same environment and computing the relative positions and orientations with Imager 5010C. Regarding the HDL-64E and Imager 5010C, the position and orientation are obtained by aligning the range data scanned by each sensor. For Ladybug 3 and Imager 5010C, the extrinsic parameter is obtained by manually specifying the correspondence point between the panorama image and the three-dimensional reflectance image and computing the extrinsic parameter using the correspondences. In {\it Scaleless} and the proposed method, eight horizontal rotation motions and eight vertical rotation motions are randomly sampled each time, and an average error of 10 times is recorded. In {\it Manual} calibration, calibration is carried out using only one scan, taking corresponding points between the panoramic image of Ladybug 3 and the 3D reflectance image of HDL-64E. For {\it MI}, MI is calculated with 16 scan sets of image-point cloud scanned 
at each position.

\begin{figure}[t]
	\begin{center}
		\includegraphics[width=\linewidth]{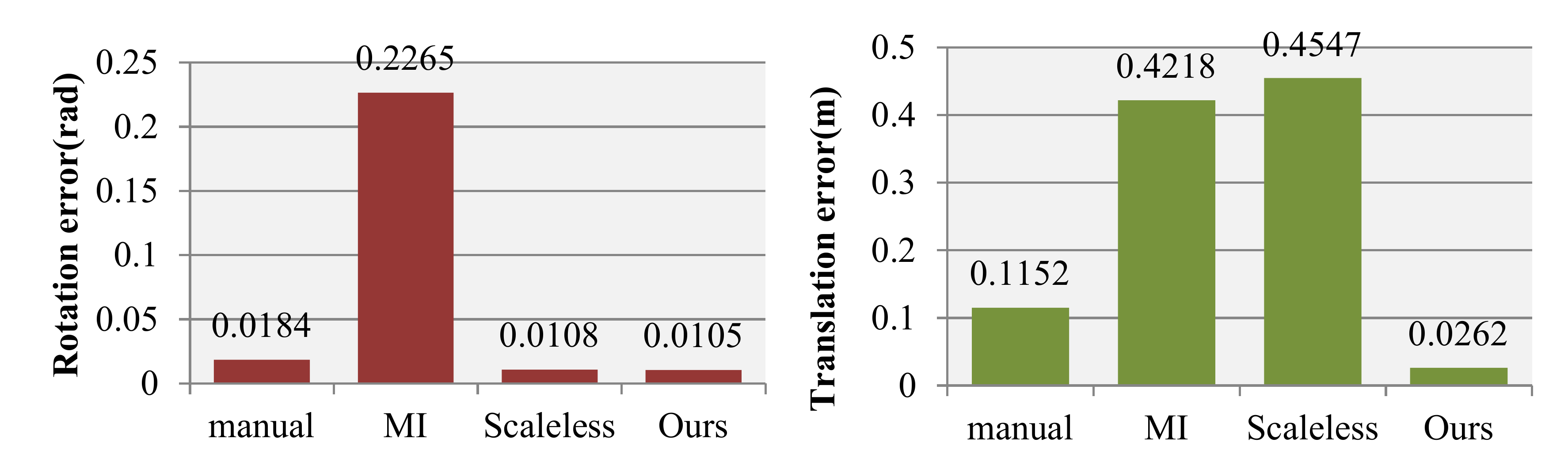}
		\caption{Error from the ground truth of the calibration result by each method using HDL-64E and Ladybug 3. In {\it MI}, {\it Scaleless}, and ours, calibration performed with 16 scans}
		\label{fig:velocomp}
	\end{center}
\end{figure}

From Fig.~\ref{fig:velocomp}, {\it Manual} calibration fails to obtain accurate results because the narrow scan range and the sparse point cloud of HDL-64E make correspondence construction difficult. Also in {\it MI}, since HDL-64E has low resolution and information of reflectance is not clear, optimization cannot be completed with this dataset. On the other hand, in the motion-based methods, accuracies are less than 1 degree in the rotation in both {\it Scaleless} and ours. However, in {\it Scaleless}, translation is significantly different from the ground truth. In contrast, in ours, highly accurate translation results are obtained.

The proposed method can also work with motions acquired by operating rover. To obtain all the extrinsic parameter of 6-DoF by hand-eye calibration based method, it is necessary to rotate in two or more directions. However, regarding the rotation motion in the vertical direction, a part of the platform on which the sensors are mounted must be raised.  Although this operation is more difficult than the horizontal rotation, this experiment demonstrates that the proposed method works well with the vertical rotational motions obtained by a reasonable mobile platform operation such as raising and lowering the small step.
%========================================================================================================
\section{Conclusion}
In this paper, we present targetless automatic 2D-3D calibration based on hand-eye calibration using sensor fusion odometry for camera motion estimation. The proposed method can be fully utilized with less translation and larger rotation camera motions. It is also preferable to carry out the measurement for calibration in the place surrounded by flat terrains as much as possible. 

It is necessary to rotate the sensor in multi directions to carry out the hand-eye calibration. However, in many cases it is more difficult to make rotation in the vertical direction than in the horizontal direction. Although the proposed method also requires satisfying this conditions, it is enough for carrying out the calibration to use vertical rotation obtained through reasonable movement by using mobile platform. Therefore this method is highly practical and it is possible to calibrate dynamically during scanning by choosing appropriate motions.

%The proposed method does not require the reflectance information or initial values of the parameters and is highly practical.

%========================================================================================================
\section*{Acknowledgment}
This work was partially supported by the social corporate program (Base Technologies for Future Robots) sponsored by NIDEC corporation and also supported by JSPS Research Fellow Grant No.16J09277.
%========================================================================================================

{\small
	\bibliographystyle{IEEEtran}

}

\end{document}